%% file: root.tex
\documentclass[letterpaper, 10pt, conference]{ieeeconf}
\IEEEoverridecommandlockouts
\input{preamble}

\input{commands}

\newcommand{\herofigure}{%
  \vspace{0.5em}%
  \begin{center}%
    \includegraphics[width=\textwidth]{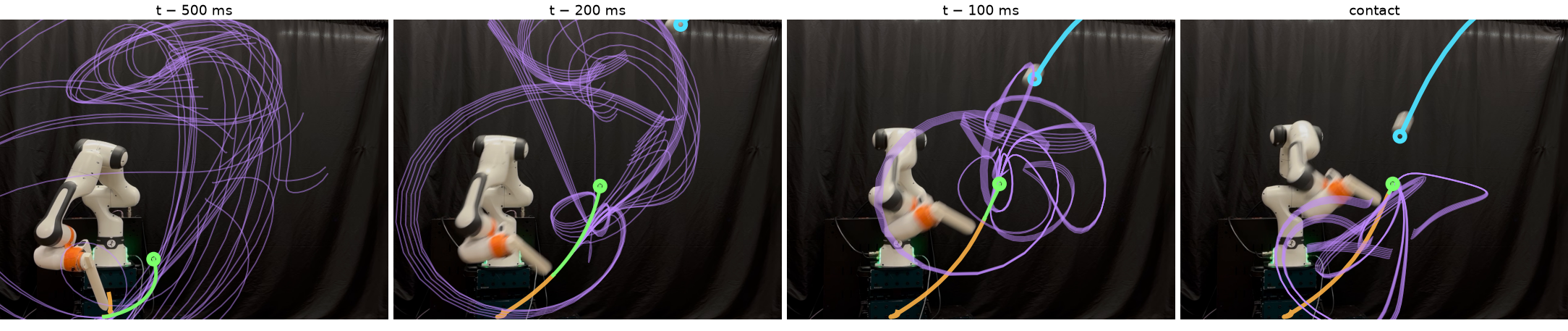}%
    \captionof{figure}{An interception sequence controlled by our planner at \(T\) minus \qtylist{500;200;100;0}{\milli\second}. Gold: end-effector history. Purple: candidate trajectories. Green: planned intercept point. Blue: object history. Our planner is capable of real-time interception under uncertainty.}%
    \label{fig:new_fig_15_hero}%
  \end{center}%
\vspace{-0.5em}}
\makeatletter
\def\maketitle{\par%
  \begingroup%
  \normalfont%
  \def\thefootnote{}%
  \def\footnotemark{}%
  \footnotesize%
  \footnotesep 0.7\baselineskip%
  \normalsize%
  \twocolumn[\@maketitle\herofigure\@IEEEdynamictitlevspace\@IEEEaftertitletext]%
  \thispagestyle{titlepagestyle}\@thanks%
  \if@IEEEusingpubid\enlargethispage{-\@pubidpullup}\fi%
  \endgroup%
  \setcounter{footnote}{0}\let\maketitle\relax\let\@maketitle\relax%
  \gdef\@thanks{}%
\let\thanks\relax}
\makeatother

\title{\fontsize{17pt}{24pt}\selectfont \bf Becoming a Fruit Ninja: Real-Time Probabilistic Kinodynamic Planning for Manipulator Projectile Interception}
\newif\ifanonymous
\anonymousfalse

\ifanonymous
\author{Anonymous Author(s)}
\else
\author{
  Lucas Chen, Austin Garrett, Andrew Niu, and Zachary Kingston%
  \thanks{LC, AG, AN, and ZK are with the Department of Computer Science, Purdue University, West Lafayette, IN, USA. {\tt \{chen4007, ajg, niua, zkingston\}@purdue.edu}.
}}
\fi

\begin{document}
\maketitle
\thispagestyle{empty}
\pagestyle{empty}

\begin{abstract}
  Projectile interception is a challenging dynamic manipulation problem.
  Intercepting a thrown object with a robot arm requires reaching a point on the object's path as the object passes through it.
  Slicing also fixes the blade's velocity and orientation at contact.
  The goal is therefore a subset of the states of the robot and arrival times that moves as the object falls, and the arm must reach it within its actuator limits in milliseconds.
  We present \methodname, an anytime sampling-based planner that grows a tree on the GPU in batches toward the interception manifold.
  Each edge is an exact cubic whose travel time is found by a parallel search against the arm's dynamics, so every edge satisfies the actuator limits.
  Plans are ranked by a risk-aware objective over the uncertainty in the object's position and the arm's arrival time.
  We evaluate on a Franka Research 3 against six baselines in a calibrated real-time simulator, where \methodname cuts \qty{96.7}{\percent} of tosses in the open and \qty{68.3}{\percent} among five obstacles, versus the best baseline's \qty{68.3}{\percent} and \qty{35.0}{\percent} respectively.
\end{abstract}

\section{Introduction}
\label{sec:intro}

Catching a thrown object is a long-standing benchmark for dynamic manipulation~\cite{birbach2011,kim2014}, and striking one, as in table tennis~\cite{durr2026}, adds a terminal velocity to the goal.
We consider slicing (\cref{fig:new_fig_15_hero}), which adds the blade's velocity and orientation at contact.
These problems couple the goal to the plan: the arm must reach the object where it will be, so the intercept point depends on the arrival time, which depends on the trajectory.
The goal is therefore a manifold of configurations, joint velocities, and arrival times.
On an arm such as the Franka Research 3 (FR3), the plan must also respect torque, power, and position-dependent velocity limits, since exceeding any one can damage the hardware.
The object is only within the robots reach for under a second and its predicted position is uncertain, so the planner has milliseconds to react and generate a trajectory.

Recent planners intercept a projectile with an industrial arm~\cite{natarajan2024,olin2025}, but start from a fixed configuration and reach a pose with zero terminal velocity.
Sampling-based motion planners (SBMPs) can solve the full problem, since kinodynamic variants~\cite{lavalle2001,webb2013} handle general goal regions and differential constraints, but interception imposes hard real-time constraints.
Two compatible advances bring SBMPs to millisecond speeds, enough for full planning inside the interception loop: GPU-parallel SBMPs~\cite{prrtc,pakr,kinopaxplus} batch tree growth into single kernels, and FLASK~\cite{flask2026} steers exactly between arm states with closed-form cubic edges.

We present \methodname, an anytime batched kinodynamic sampling-based motion planner for real-time interception of airborne objects.
At \qty{50}{\hertz}, it samples batches of candidate states, both uniformly and from an explicit nine-dimensional chart of the goal manifold over arrival time, contact point, blade orientation, and speed, and connects each to its nearest tree nodes by a minimum-time cubic.
Each edge is timed by a parallel search against a dynamics model of the FR3, and edges that violate its limits are discarded.
An anytime outer loop keeps the incumbent plan and prunes nodes that cannot reach the object in time, and a \qty{1}{\kilo\hertz} controller refits a cubic to the planner's rendezvous every cycle.
Since the arrival point is an estimate, plans are ranked by a risk-aware objective that trades cut speed against the probability of contact under uncertainty that grows with the arrival time.
We evaluate on hardware and in a simulator calibrated on the physical arm, against six baselines.
\methodname cuts \qty{96.7}{\percent} of tosses in the open and \qty{68.3}{\percent} among five obstacles, against \qty{68.3}{\percent} and \qty{35.0}{\percent} for the best baseline, and ablations show that the cubic edge and the dynamics model are both necessary.

\section{Related Work}
\label{sec:related}

\citet{croft1998} pose interception as selecting the earliest rendezvous point on the object's predicted path that the arm can reach, which couples the goal to the travel time.
Catching systems pair a filter over the object's flight~\cite{birbach2011} with online optimization over catch posture and timing~\cite{kim2014}.
Slicing is closer to striking tasks such as table tennis~\cite{durr2026}, where the terminal velocity is part of the goal.
Closest to ours, \citet{natarajan2024} precompute a library of trajectories from a home configuration to a discretized set of shield poses, and \citet{olin2025} precompute a tree of jerk-limited primitives over a similar goal set and select a branch online as the belief over the crossing point tightens.
Both block the projectile with an industrial arm, plan to a pose with zero terminal velocity, and enforce only kinematic limits.
Our planner replans every cycle, checks each edge against the arm's dynamics, and plans to a goal that includes velocity and orientation at contact.

Kinodynamic SBMPs propagate sampled controls when no steering function is available~\cite{lavalle2001}, or steer exactly when the boundary value problem has a closed form~\cite{webb2013}.
FLASK~\cite{flask2026} solves it for manipulator arms through differential flatness, where the solutions are cubics.
Asymptotic optimality comes from rewiring~\cite{karaman2011}, dominance pruning~\cite{li2016}, or AO-x~\cite{hauser2016}, which wraps a feasible planner in a cost-bounded outer loop and needs no cost-to-go.
These trees grow one node at a time, which limits throughput in a real-time setting.
GPU parallelism now covers geometric SBMPs~\cite{prrtc}, manipulation~\cite{spasm,cutamp}, and trajectory optimization~\cite{curobo2023}.
In the kinodynamic setting, Kino-PAX\(^+\)~\cite{kinopaxplus} decomposes tree growth into independent GPU subroutines and PAKR~\cite{pakr} vectorizes a kinodynamic RRT so that a batch of propagations is one kernel.
Our planner follows the batched inner loop of PAKR within an anytime outer loop, but steers with cubics timed by a parallel minimum-time search.

Planning against a target that moves while the planner runs is a replanning problem.
RRT\(^X\)~\cite{otte2016} maintains a single graph rewired online as the world changes, and \citet{hauser2012} gives conditions under which such a loop stays safe and complete when every cycle has a hard time bound.
\citet{covic2025} treat each replanning cycle as a periodic real-time task, and KRAFT~\cite{kraft2024} pairs a kinodynamic tree planner with feedback tracking and a contingency stop.
Trajectory optimization~\cite{trajopt2014,curobo2023} and sampling-based MPC~\cite{mppi2017,bhardwaj2021} refine a trajectory locally over a fixed horizon, so they converge to local minima and do not plan ahead to a rendezvous.
Online trajectory parameterization approaches~\cite{kroger2010,berscheid2021} emit a limit-respecting reference to a given target state every cycle, but do not choose the target.
Replanning from scratch each cycle solves kinodynamic problems that a single query cannot~\cite{tsianos2008,sabbadini2026replan}.
Our planner rebuilds the tree from the measured state every planning cycle, since the goal moves between cycles, and sends only the first segment of the plan to the controller.

The arrival point is an estimate, so plans must be ranked under uncertainty.
Chance-constrained planning bounds the probability of collision along a path, in sampling-based planners~\cite{luders2010} and in trajectory optimization~\cite{blackmore2011,brudermuller2026cc}, and risk-aware planners derive that bound from reachable sets~\cite{liu2023radius}.
These methods treat uncertainty as a constraint to satisfy, while \citet{long2013} propagate it to choose an intercept point for a planar target.
Our objective scores the probability of contact under uncertainty that grows with the chosen arrival time, and ranks plans by it.

\section{Problem Statement}
\label{sec:problem}

Let \(x : \R_{\ge 0} \to \R^3\) be the object's flight, estimated online from its initial position \(x_0\) and velocity \(v_0\) under the ballistic model
\begin{equation}
  x(t) = x_0 + v_0 t + \tfrac{1}{2} g\, t^2.
  \label{eq:ballistic}
\end{equation}
The object is within the arm's reach for \(t \in [t_{\mathrm{enter}}, \tfall]\), a window of under a second.

The arm has configuration \(q \in \mathcal{C} \subset \R^7\) and state \((q, \dot q)\), $\dot q \in \R^7$.
A trajectory \(\sigma : [0, T] \to \mathcal{C}\) over travel time \(T\) requires joint torques \(\tau(t)\) given by the arm's inverse dynamics (\cref{sec:dynamics}).
The FR3 bounds each joint's position, velocity, and torque and the total mechanical power, $W$, where the velocity bound \(\bar{\dot q}_i(q_i)\) tapers with position.

\begin{definition}[Feasible trajectory]\label{def:feasible}
  A trajectory \(\sigma\) is feasible if it is collision-free and, \(\forall t \in [0,T]\) and joints \(i\),
  \begin{equation*}
    \begin{gathered}
      |\dot\sigma_i(t)| \le \bar{\dot q}_i(\sigma_i(t)), \qquad
      |\tau_i(t)| \le \tau_i^{\max},\\
      \Big|\sum_i \tau_i(t)\,\dot\sigma_i(t)\Big| \le W^{\max}.
    \end{gathered}
  \end{equation*}
\end{definition}

Exceeding a bound triggers one of the FR3's velocity, power, or torque-rate reflexes, which decelerate the arm without a fault, so feasibility is a hard constraint.
For a fixed boundary state, the required torque and power reduce as \(T\) grows, which \cref{sec:mintime} uses; the velocity taper does not, since it depends on position.

The end effector carries a blade with edge \(\mathcal{B}(q) \subset \R^3\) from forward kinematics.
For a contact point \(p\), let \(v_b(q, \dot q, p)\) be the blade's velocity there and \(n_b(q)\) its cutting direction.

\begin{definition}[Interception manifold]\label{def:manifold}
  The interception manifold is the set of arm states and arrival times at which the blade cuts the object,
  \begin{equation*}
    \begin{aligned}
      \Mman = \big\{\, (q, \dot q, T) \;\big|\;\; & x(T) \in \mathcal{B}(q),\;\; a(q, \dot q, T) \ge a_{\min}, \\
      & \|v_b(q, \dot q, x(T)) - \dot x(T)\| \ge v_{\min} \,\big\},
    \end{aligned}
  \end{equation*}
  where \(a(q, \dot q, T) = \cos\angle\big(n_b(q),\, v_b - \dot x(T)\big)\) is the alignment between the cutting direction and the relative velocity, and \(a_{\min}\) and \(v_{\min}\) are the alignment and cut-speed thresholds.
\end{definition}

Our experiments use \(a_{\min} = \num{0.94}\) and \(v_{\min} = 3.0\,\unit{\metre\per\second}\).
The manifold moves with \(T\) as the object moves.
Since \(x(\cdot)\) is an estimate whose error grows with \(T\), membership in \(\Mman\) is uncertain, and \cref{sec:objective} ranks candidates by probability of contact.

\begin{definition}[Interception planning]\label{def:problem}
  Given the arm's measured state \((q_0, \dot q_0)\) and the estimated flight \(x(\cdot)\), find an arrival time \(T \in [t_{\mathrm{enter}}, \tfall]\) and a feasible trajectory \(\sigma : [0,T] \to \mathcal{C}\) such that
  \begin{equation*}
    (\sigma(0), \dot\sigma(0)) = (q_0, \dot q_0), \qquad
    (\sigma(T), \dot\sigma(T), T) \in \Mman,
  \end{equation*}
  maximizing the objective \(J\) of \cref{eq:objective}.
\end{definition}

The goal is coupled to the plan: the intercept point \(x(T)\) depends on \(T\), and the least \(T\) for which a feasible \(\sigma\) exists depends on where \(x(T)\) is.

\section{Method}

We present \methodname (\textbf{F}ree-terminal-time \textbf{R}endezvous, \textbf{U}ncertainty-aware \textbf{I}nterception \textbf{T}ree, a\textbf{N}ytime, \textbf{IN}cumbent-pruned, \textbf{J}AX-\textbf{A}ccelerated), a sampling-based planner that solves \cref{def:problem} from the measured state at \qty{50}{\hertz}. Each batch samples the goal manifold through an explicit chart and free states near the chord to it, connects every sample to its nearest tree nodes by a minimum-time cubic, and keeps the edges that satisfy the arm's actuator limits and clearance.
Samples on the manifold are scored by a risk-aware objective over the uncertainty in the object's flight.
An anytime outer loop keeps the best rendezvous found so far, prunes nodes that cannot reach the object before it, and sends the first edge of that plan to the controller when the cycle's time-limit is exceeded.
We describe the chart, the objective, the dynamics model, the cubic edge, the planner, and the controller in turn.

\subsection{The goal chart}
\label{sec:goalmanifold}

\cref{def:manifold} defines \(\Mman\) implicitly. %
Given the flight \(x(\cdot)\), however, a rendezvous is fixed by a small number of choices, so \(\Mman\) has an explicit chart and can be sampled directly.

\begin{definition}[Goal chart]\label{def:chart}
  A chart of \(\Mman\) is a pair \((\mathcal{Z}, \phi)\) of a parameter space \(\mathcal{Z} = [0,1]^9\) and a decoding map to configuration, velocity, and arrival time \(\phi : \mathcal{Z} \to \mathcal{C} \times \R^7 \times \R_{\ge 0}\), defined on a subset of \(\mathcal{Z}\), such that \(\phi(z) = (q_g, \dot q_g, T) \in \Mman\) wherever it is defined.
\end{definition}

The $\phi(z)$ map is defined as follows.
The first six coordinates of $z$ set the intercept: \(z_0\) selects the arrival time \(T \in [t_{\mathrm{enter}}, \tfall]\) and thus the contact point \(x(T)\) via~\eqref{eq:ballistic}; \(z_1, z_2\) set the blade's approach direction as a lateral offset within a cone about \(-\dot x(T)\); and \(z_3, z_4, z_5\) set the blade's spin, roll, and tilt, completing the end-effector rotation \(R_g \in SO(3)\).
The remaining three resolve redundancy: \(z_6\) selects the FR3's wrist angle in \([q_7^{\min}, q_7^{\max}]\), \(z_7\) picks the contact point \(p \in \mathcal{B}(q_g)\) along the \qty{30}{\centi\metre} edge, and \(z_8\) adds closing speed beyond the object's own velocity.
Together these fix the end-effector pose \((p_g, R_g)\), from which closed-form inverse kinematics~\cite{HeLiu2021} gives \(q_g\), and the terminal joint velocity follows as \(\dot q_g = J^\dagger(q_g)\, v_{\mathrm{ee}}\), where \(v_{\mathrm{ee}}\) combines \(\dot x(T)\) and the closing speed.
The approach cone and closing speed enforce the alignment and speed conditions of \cref{def:manifold} by construction.
When inverse kinematics has no solution or \(\dot q_g\) exceeds bounds, \(\phi\) is undefined and the sample is discarded.

\subsection{Objective}
\label{sec:objective}

The flight estimate has independent position, velocity, and acceleration uncertainty \(\sigma_p\), \(\sigma_v\), and \(\sigma_a\), which propagate to a position uncertainty at time \(t\) of
\begin{equation*}
  \sigma^2(t) = \sigma_p^2 + (\sigma_v t)^2 + (\tfrac{1}{2}\sigma_a t^2)^2.
\end{equation*}
The score of a rendezvous \((q_g, \dot q_g, T) \in \Mman\) is
\begin{equation}
  \begin{aligned}
    J &= w_v \cdot \min(v_{\mathrm{rel}},\; v_{\mathrm{target}}) \\
    &+ w_p \cdot \log P_{\mathrm{object}} + w_r \cdot \log P_{\mathrm{inrange}} + w_x \cdot \log P_{\mathrm{arm}},
  \end{aligned}
  \label{eq:objective}
\end{equation}

with
\vspace{-1em}

\begin{equation}
  \begin{aligned}
    v_{\mathrm{rel}} &= \|v_b - \dot x(T)\|, \\
    P_{\mathrm{object}} &= \operatorname{erf}\!\big(r / \sigma(T)\sqrt{2}\big), \\
    P_{\mathrm{inrange}} &= \Phi\!\big((\tfall - T) / \sigma_t\big), \quad \sigma_t = \sigma(T) / |\dot x_z(\tfall)|, \\
    \log P_{\mathrm{arm}} &= -\lambda(\nu - \nu_0)_+\, T\, (1 - \nu_0/\nu).
  \end{aligned}
  \label{eq:terms}
\end{equation}
The first term rewards the relative speed at contact, clipped at \(v_{\mathrm{target}}\).
\(P_{\mathrm{object}}\) is the probability that the object lies within the catch radius \(r\) of \(x(T)\), and \(P_{\mathrm{inrange}}\), with \(\Phi\) the standard normal cumulative distribution function, is the probability that the object has not yet left reach.
\(P_{\mathrm{arm}}\) penalizes a peak joint-velocity fraction \(\nu\) above the reflex threshold \(\nu_0\), in proportion to the time spent above it.
Because \(\sigma(T)\) grows with the arrival time, the score trades cut speed against arriving before the estimate's uncertainty grows.

\subsection{Dynamics model}
\label{sec:dynamics}

Both the planner's feasibility check and the \qty{1}{\kilo\hertz} controller require the joint torque needed to follow a given trajectory.
For a state \((q, \dot q, \ddot q)\) along an edge, the required torque at joint \(i\) is
\begin{equation}
  \tau_i = \big[\,M(q)\,\ddot q + c(q,\dot q) + g(q)\,\big]_i + I_{m,i}\,\ddot q_i + \tau_{f,i}(\dot q_i),
  \label{eq:torque}
\end{equation}
where \(M(q)\ddot q + c(q,\dot q) + g(q)\) is the rigid-body inverse dynamics (recursive Newton--Euler, RNEA), \(I_{m,i}\) is the reflected rotor inertia of joint \(i\), and \(\tau_{f,i}(\dot q_i)\) is a friction model.
We adopt the identification of \citet{gaz2019}, who fit both terms on the Panda arm, the FR3's mechanical predecessor.

The reflected rotor inertia \(I_{m,i} = J_{m,i} \cdot n_i^2\) (motor inertia times squared gear ratio) is the torque needed to accelerate the rotor itself.
RNEA models only the link-side bodies, and omitting this term under-predicts the torque required for fast motions by up to \qty{12}{\percent} of peak torque on the proximal joints.
The friction model is a per-joint sigmoid~\cite{gaz2019},
\begin{equation}
  \tau_{f,i}(\dot q_i) = \frac{\psi_{1,i}}{1 + e^{-\psi_{2,i}(\dot q_i + \psi_{3,i})}} - \frac{\psi_{1,i}}{1 + e^{-\psi_{2,i}\,\psi_{3,i}}},
  \label{eq:friction}
\end{equation}
where \(\psi_{1,i}\) sets the saturation amplitude, \(\psi_{2,i}\) the transition sharpness, and \(\psi_{3,i}\) a velocity offset, and the second term enforces \(\tau_{f,i}(0) = 0\). We choose the same values for these constants as those provided by Gaz et al.
The FR3 firmware adds friction compensation to commanded torques, and its safety reflexes are checked against the arm's internal torque estimates, so the friction model is needed to predict and avoid reflex triggers.

\subsection{Cubic edges and minimum-time search}
\label{sec:mintime}

FLASK~\cite{flask2026} plans in the flat output space of a differentially flat system, where the boundary value problem has a closed-form solution.
A fully actuated arm is flat in its joint positions, \(y = q\), and with joint acceleration as the pseudo-control, the minimum-time, minimum-effort trajectory between two boundary states is a per-joint cubic. We refer the reader to the original paper for details on the construction of the cubic.

\begin{definition}[Cubic edge]\label{def:edge}
  The edge from \((q_0, \dot q_0)\) to \((q_1, \dot q_1)\) with travel time \(T\) is
  \begin{equation}
    \sigma_T(t) = q_0 + \dot q_0\, t + \Big(\tfrac{3}{T^2} d_1 - \tfrac{1}{T} d_2\Big) t^2 + \Big(\tfrac{1}{T^2} d_2 - \tfrac{2}{T^3} d_1\Big) t^3,
    \label{eq:cubic}
  \end{equation}
  with \(d_1 = q_1 - q_0 - \dot q_0 T\) and \(d_2 = \dot q_1 - \dot q_0\).
\end{definition}

The travel time fixes the edge's shape as well as its speed, so retiming a cubic changes its path, unlike time-optimal path parameterization~\cite{bobrow1985}, which retimes a fixed path.
FLASK chooses \(T\) by minimizing a time-plus-effort cost and validates the edge against bounds on joint acceleration.
The FR3's limits are on torque and power, which depend on configuration and velocity through~\eqref{eq:torque}, and valid FLASK cubics often violate these terms.
Discarding such edges leaves the tree with few feasible connections, so we instead search for \(T\) checked against the dynamics model, as discussed below.

\begin{definition}[Feasibility Check]\label{def:check}
  Evaluate \(\sigma_T\) at \(S\) equally spaced times \(t_s\), with \(\tau(t_s)\) the torque of~\eqref{eq:torque}:
  \begin{equation}
    \begin{aligned}
      \rho(T) &= \max\big(\rho_\tau,\; \rho_P,\; \rho_v\big), \quad
      \rho_\tau = \max_{i,s} \frac{|\tau_i(t_s)|}{\tau_i^{\max}}, \\
      \rho_P &= \max_s \frac{\big|\sum_i \tau_i(t_s)\,\dot\sigma_{T,i}(t_s)\big|}{W^{\max}}, \quad
      \rho_v = \max_i \frac{|\dot\sigma_{T,i}|_{\mathrm{peak}}}{\bar{\dot q}_i(\sigma_{T,i})},
    \end{aligned}
    \label{eq:rho}
  \end{equation}
  and the edge is feasible iff \(\rho(T) \le 1\).
\end{definition}

The three terms check joint torque, mechanical power (\(W^{\max} = \qty{120}{\watt}\)), and the position-dependent velocity taper.
The velocity term uses the exact extremum of the cubic's quadratic velocity profile rather than the samples.
For fixed boundary states, \(\rho_\tau\) and \(\rho_P\) decrease monotonically in \(T\), since a slower motion requires less torque and power, whereas \(\rho_v\) depends on position alone and does not.

\begin{definition}[Minimum feasible time]\label{def:mintime}
  The minimum feasible time of an edge is
  \begin{equation*}
    T^\star = \min\,\{\, T > 0 \mid \rho_\tau(T) \le 1 \text{ and } \rho_P(T) \le 1 \,\}.
  \end{equation*}
\end{definition}

We find \(T^\star\) by search; a binary search underuses a GPU, so we use an \(N\)-ary search: the interval \([T_{\mathrm{lo}}, T_{\mathrm{hi}}]\) is split into \(N\) equal slots, all \(N\) are evaluated in parallel, and the single feasible sub-interval of width \((T_{\mathrm{hi}} - T_{\mathrm{lo}})/N\) is refined on the next iteration.
This converges in \(\lceil \log_N(T_{\mathrm{hi}} / \epsilon) \rceil\) rounds rather than \(\lceil \log_2(T_{\mathrm{hi}} / \epsilon) \rceil\), at the cost of \(N\) evaluations per round.
Since \(\rho_v\) is not monotonic in \(T\), it is checked once at \(T^\star\), and edges that violate it are discarded.

\subsection{The \methodname planner}
\label{sec:planner}

\methodname (\cref{alg:abmp}) grows a tree of cubic edges (\cref{def:edge}) from the measured state.
The tree is expanded in batches on the GPU and stored in a preallocated struct-of-arrays layout.
An anytime outer loop repeats the expansion and keeps the rendezvous node with the highest score \(J\) of~\eqref{eq:objective}.
The tree \(\mathcal{T}\) is a set of nodes \(n = (q_n, \dot q_n, t_n)\), each joined to its parent by a feasible cubic edge, where \(t_n\) is the time-to-come along the edges from the root \((q_0, \dot q_0, 0)\).
A node is a rendezvous node if \((q_n, \dot q_n, t_n) \in \Mman\).

\emph{Collision checking.}
The arm's collision geometry is a set of spheres covering each link.
Self-collision pairs that the joint limits make unreachable are removed offline.
An edge is checked by a discrete set of configurations along its cubic and computing the minimum clearance between the arm spheres and the obstacle capsules at each sample.
An edge whose minimum clearance is below a safety margin is rejected.
Because the check is per edge, a goal whose direct cubic is in collision can still be reached through an intermediate node, which a single-cubic planner cannot do.
During the search, clearance is checked at a coarse resolution, and the plan sent to the controller is rechecked at full resolution, since checking every candidate edge at full resolution would be the largest cost in the cycle.

\emph{Batched expansion.}
Each round samples \(B\) candidate states.
A fraction \(p_{\mathrm{goal}}\) of them are goal candidates: parameters drawn from a density \(p_\theta\) over the chart and decoded by \(\phi\) (\cref{ln:sample}).
Each goal candidate has a required arrival time \(T_{\mathrm{req}}\), set by its \(z_0\).
These samples bias growth toward the reachable part of \(\Mman\).
The density \(p_\theta\) is a Gaussian mixture over \(\mathcal{Z}\) mixed with a uniform component.
It is refit between anytime iterations to the parameters that produced the best plans, and it persists across planning cycles.
The mixture concentrates samples near previously found rendezvous; the uniform component keeps every point of the chart reachable.
The remaining candidates are free states (\cref{ln:free}).
The configuration of a free state is a Gaussian perturbation of a point on the chord in joint space from the root to a decoded goal, again mixed with a uniform component over the joint box.
Its velocity is drawn from a fraction of the joint-velocity limits at that configuration.

Each candidate has \(M\) parent candidates: the root and its \(M-1\) nearest tree nodes (\cref{ln:parents}), found by a brute-force parallel scan.
The root is always included because it could potentially find cheaper and smoother paths in early iterations, in comparison to connecting with the nearest neighbor set, which produces segmented paths that are jerky and require more iterations to smooth out.

A goal candidate must arrive at \(T_{\mathrm{req}}\), so its edge duration from parent \(p\) is fixed at \(T_{\mathrm{req}} - t_p\) and the check is evaluated once (\cref{ln:goaldur}).
A free candidate has no required arrival time, but is still bounded by \(T_{\mathrm{best}}\). To efficiently compute an admissible bound for this time, we rely on one iteration of an \(N\)-ary search.
Edges that fail the check of \cref{def:check} or the clearance check are discarded (\cref{ln:clear}).
Each candidate is attached to the remaining parent that gives it the smallest time-to-come (\cref{ln:attach}).
Multiple parents are needed for detours: if the edge from the nearest node is in collision, the edge from another parent may be clear.

\emph{Anytime outer loop.}
After every round, the incumbent is set to the highest-scoring rendezvous node in the tree (\cref{ln:incumbent}).
At the end of each anytime iteration, the tree is pruned (\cref{ln:prune}).
Informed pruning needs a cost-to-go bound, which has no closed form here.
We instead prune on arrival time.
Let \(\sigma^\star = \argmax_{c} J(c)\) over the rendezvous nodes of \(\mathcal{T}\) and \(t_{\mathrm{best}} = t_{\sigma^\star}\).
For a node \(n\) with blade position \(p_n\), let
\begin{equation}
  T_n = \min\,\Big\{\, T \in [t_{\mathrm{enter}}, \tfall] \;\Big|\; t_n + \frac{\|p_n - x(T)\|}{V_{\mathrm{tip}}} \le T \,\Big\}
  \label{eq:reach}
\end{equation}
be the earliest time at which the blade can reach the object from \(n\), where \(V_{\mathrm{tip}}\) bounds the end-effector speed.
The node is pruned if \(T_n\) does not exist or \(T_n > t_{\mathrm{best}}\).

{ %

\begin{algorithm}[t]
  \caption{\methodname: one planning cycle}
  \label{alg:abmp}
  \small
  \SetKwFunction{Chord}{ChordSample}
  \SetKwFunction{Parents}{Parents}
  \SetKwFunction{Clear}{Clear}
  \SetKwFunction{Prune}{Prune}
  \SetKwInOut{Input}{Input}
  \SetKwInOut{Output}{Output}
  \DontPrintSemicolon
  \Input{root \((q_0,\dot q_0)\); flight \(x(\cdot)\); batch \(B\);
  parents \(M\); rounds \(R\); anytime iterations \(A\)}
  \Output{rendezvous \((q_g,\dot q_g,T)\) for the \qty{1}{\kilo\hertz} controller}
  \(\mathcal{T} \gets \{(q_0,\dot q_0,\,0)\}\), \(\sigma^\star \gets \varnothing\), \(t_{\mathrm{best}} \gets \tfall\)\;
  \For{\(a = 1 \dots A\)}{
    \For{\(r = 1 \dots R\)}{
      \(Z \sim p_\theta\), \(|Z| = p_{\mathrm{goal}} B\)\label{ln:sample}\;
      \(X_{\mathrm{goal}} \gets \phi(Z, x(\cdot))\)\;
      \(X_{\mathrm{free}} \gets\) \Chord{\(q_0, X_{\mathrm{goal}}, (1{-}p_{\mathrm{goal}})B\)}\label{ln:free}\;
      \ForEach{\(x \in X_{\mathrm{goal}} \cup X_{\mathrm{free}}\) \textbf{\textup{in parallel}}}{
        \(P \gets\) \Parents{\(\mathcal{T}, x, M\)}\label{ln:parents}\;
        \ForEach{\(p \in P\) \textbf{\textup{in parallel}}}{
          \eIf{\(x \in X_{\mathrm{goal}}\)}{
            \(T_p \gets T_{\mathrm{req}}(x) - t(p)\) \(\mathrm{ok}_p \gets \rho(T_p) \le 1\)\label{ln:goaldur}\;
          }{
            \(T_p \gets \min\{T \le t_{\mathrm{best}} \mid \rho(T) \le 1\}\)\;
            \(\mathrm{ok}_p \gets T_p < \infty\)\;
          }
          \(\mathrm{ok}_p \gets \mathrm{ok}_p \wedge\) \Clear{\(p, x, T_p\)}\label{ln:clear}\;
        }
        \If{\(\exists\, p \in P : \mathrm{ok}_p\)}{
          \(p^\star \gets \argmin_{p \,:\, \mathrm{ok}_p} t(p) + T_p\)\;
          \(\mathcal{T} \gets \mathcal{T} \cup \{(x, p^\star, t(p^\star) + T_{p^\star})\}\)\label{ln:attach}\;
        }
      }
      \If{\(\mathcal{T} \cap \Mman \neq \varnothing\)}{
        \(\sigma^\star \gets \argmax_{c \,\in\, \mathcal{T} \cap \Mman} J(c)\)\label{ln:incumbent}\;
        \(t_{\mathrm{best}} \gets t(\sigma^\star)\)\;
      }
    }
    \(\mathcal{T} \gets\) \Prune{\(\mathcal{T}, t_{\mathrm{best}}\)}\label{ln:prune}\;
  }
  \Return first segment of \(\sigma^\star\)\;
\end{algorithm}

} %

\subsection{Controller and execution architecture}
\label{sec:controller}

\begin{figure}[t]
  \centering
  \includegraphics[width=\linewidth]{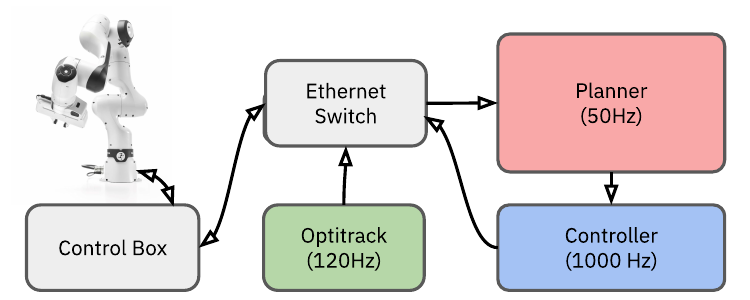}
  \caption{System architecture. The planner observes the object state and produces a commanded reference \((q_g, \dot q_g, T)\) at \qty{50}{\hertz}; the controller refits a cubic from measured state and commands torques at \qty{1}{\kilo\hertz}. The planner's latency compensation shifts the target forward by \(\Delta t_{\text{plan}} \approx \qty{20}{\milli\second}\).}
  \label{fig:timing}

  \vspace{-1em}

\end{figure}

The system operates at two rates (\cref{fig:timing}).
An incumbent plan is a path of one or more cubic edges, the planner only uses the endpoint of the first, which is sent to the \qty{1}{\kilo\hertz} tracking controller \((q_g, \dot q_g, T)\), and the tree is rebuilt at the next planning cycle.
Each control cycle, the controller fits its own cubic from the measured state \((q, \dot q)\) to the rendezvous over horizon \(T\), extracts the desired acceleration
\begin{equation}
  \ddot q_{\text{des}} = \text{clip}\!\Big(\frac{6(q_g - q)}{T^2} - \frac{4\dot q + 2\dot q_g}{T},\; [\ell, u]\Big),
  \label{eq:controller}
\end{equation}
where \([\ell, u]\) taper the velocity toward zero near the joint limits, and applies the dynamics model~\eqref{eq:torque} to obtain a torque command \(\tau \in \R^7\), clipped to the FR3's torque, slew, and power limits.
Because the refit starts from the measured state every cycle, the controller needs only the current rendezvous and no knowledge of the tree. The controller runs at \qty{1}{\kilo\hertz} in step with the FR3's required control rate.

The planner runs at \qty{50}{\hertz} on a separate thread. Unlike the controller, this requires compensation for its own latency: a plan produced at time \(t\) targets the predicted object state at \(t + \Delta t_{\text{plan}}\), where \(\Delta t_{\text{plan}} \approx \qty{20}{\milli\second}\) is the measured planning time.
Without this compensation, the trajectory no longer intersects the object and the success rate drops to nearly \qty{0}{\percent}. 
Our GPU-based algorithm makes use of a unified kernel with static memory requirements, producing a highly predictable runtime with only \(\pm \qty{1}{\milli\second}\) of range.

\section{Experiments}

We evaluate \methodname against six baselines in simulation, in a clear workspace and under clutter, ablate its design choices, and validate on hardware.
All simulated experiments run on an NVIDIA RTX 4090 GPU with an AMD Threadripper CPU.
\methodname is implemented in JAX~\cite{jax2018}.

For all experiments, we fix the constants described throughout the different stages. For the scoring function, we set \(w_v = \num{0.20}\) with \(v_{\mathrm{target}} = \qty{9.0}{\metre\per\second}\), \(w_p = 8\), \(w_r = 8\), \(w_x = 1\), \(r = \qty{9}{\centi\metre}\), \(\nu_0 = \num{0.90}\) of the tapered velocity limit, and \(\lambda = \num{1.2}\).
In simulation the covariance uses the estimator's measured residuals, \(\sigma_p = \qty{4.8}{\milli\metre}\) and \(\sigma_v = \qty{18.6}{\milli\metre\per\second}\) with \(\sigma_a = 0\).
On hardware the same estimator is calibrated to \(\sigma_p = \qty{5.4}{\milli\metre}\), \(\sigma_v = \qty{40.6}{\milli\metre\per\second}\), and \(\sigma_a = \qty{0.20}{\metre\per\second\squared}\), where the nonzero \(\sigma_a\) is the unmodelled gravity tilt of the object's flight.
For the planning iterations and batch sizes, we use \(B = 128\) candidates and \(M = 4\) parents, with \(R = 2\) rounds per anytime iteration and \(A = 2\) iterations per cycle.

\input{big_block_of_tables}

\subsection{Methodology}
\label{sec:setup}

The simulator runs the same planner, controller, and communication stack as the physical system, as separate processes over UDP; only the physics and perception are simulated.
Every noise source is fitted from hardware recordings.
The simulator synthesises raw marker detections and feeds them through the estimator that runs on hardware, applies a correlated disturbance torque to the arm calibrated so that tracking divergence matches hardware replay to within \qty{8}{\percent}.
The arm uses a torque and safety model fitted from \num{300} total real swings.

Each evaluation is \num{180} tosses across three seeds.
Tosses are ballistic arcs that sweep laterally across the arm's reachable workspace with transit times of \qtyrange{0.75}{0.95}{\second}, drawn from three launch azimuths and two launch radii.

We report \emph{cut rate}, the percentage of tosses where the blade contacts the object with alignment and speed above the thresholds of \cref{def:manifold}; \emph{catch rate}, where the blade contacts the object at all; \emph{cut speed} \(\bar v\), the mean relative normal blade--object speed at contact; and, under clutter, \emph{thru}, the fraction of plans whose executed motion passes through an obstacle.
Cut rate is the primary metric; the gap to catch rate separates tosses that miss the object from those that reach it without cutting.

The baselines are \emph{CEM}~\cite{deboer2005} and \emph{CMA-ES}~\cite{hansen2001}, which sample and refine collocations points and intercept configurations in the rendezvous space; \emph{TrajOpt}~\cite{trajopt2014}, sequential convex optimization over a collocated trajectory; \emph{MPPI}~\cite{mppi2017}, which samples control sequences and takes their exponentially weighted mean; \emph{cuRobo}~\cite{curobo2023}, a GPU planner over B-spline trajectories; and \emph{RRTC}~\cite{flask2026}, a CPU bidirectional RRT with the same cubic steering as \methodname, which separates the effect of the tree from that of the GPU.
Every baseline output is converted to the controller's native format \((T, q_g, \dot q_g)\), a single cubic to the rendezvous.

\subsection{Interception in the open}
\label{sec:open}

\cref{fig:new_fig_16_frontier} plots cut rate against median solve time for every configuration of every planner.
\methodname cuts \qty{96.7}{\percent} of tosses at a median solve time of \qty{12}{\milli\second}, inside the \qty{20}{\milli\second} planning cycle.
No single-cubic baseline exceeds \qty{70}{\percent} at any solve time (\(N=0\) in \cref{tab:clutter}).
RRTC catches as many tosses as \methodname (\qty{98.3}{\percent}) but cuts only \qty{73.3}{\percent}, at \qty{4.76}{\metre\per\second} against \qty{5.35}{\metre\per\second}: the CPU tree lacks the time budget to give a solution on every solve, leaving the controller executing stale rendezvous.
CEM misses the object on a quarter of tosses, indicating that naive search over the collocation and intercept space is too slow for realtime use.

\subsection{Clutter}
\label{sec:clutter}

\begin{figure*}[ht]
  \centering
  \makebox[\linewidth][c]{\includegraphics[width=\linewidth]{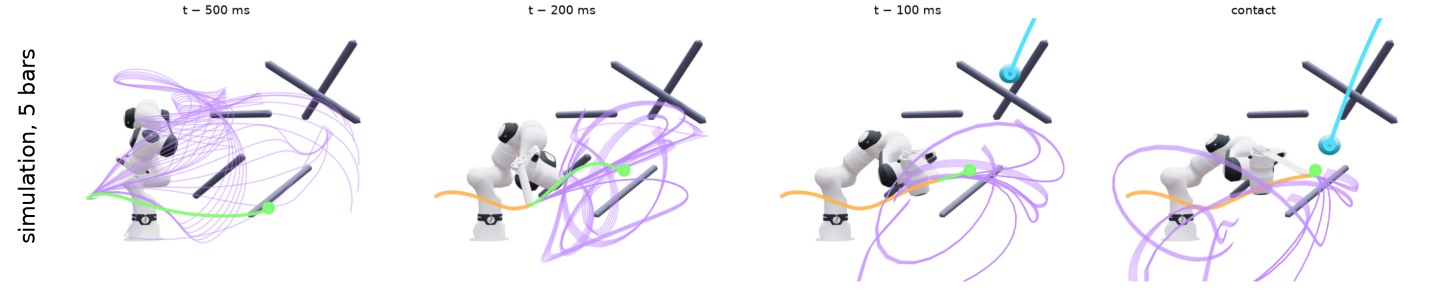}}
  \vspace{-2em}
  \caption{An example of a simulated intercept sequence with obstacles. We insert up to five 750mm long capsule shaped obstacles into the workspace of the FR3 and benchmark planning for each planner. The results for 0, 3, and 5 capsules are reported in \cref{tab:clutter}.}
  \label{fig:sim}
\end{figure*}

\cref{tab:clutter} adds three and five \qty{750}{\milli\metre} capsule obstacles to the workspace, shown in~\cref{fig:sim}, with every planner at its best open-workspace configuration and any obstacle-aware option enabled.
\methodname falls from \qty{96.7}{\percent} to \qty{81.7}{\percent} and \qty{68.3}{\percent}; the best baseline, CMA-ES, falls from \qty{68.3}{\percent} to \qty{65.0}{\percent} and \qty{35.0}{\percent}.
The obstacles illustrate the power of search, so direct optimization methods become stuck once the direct path is blocked, whereas \methodname routes through intermediate waypoints and checks each edge for collision (\cref{sec:planner}).
RRTC also builds an obstacle avoiding tree but falls to \qty{21.7}{\percent} at five capsules, because it often fails to find a solution within the time budget.
MPPI never plans ahead to a rendezvous and cuts nothing; cuRobo is too slow to finish a run even at its fastest setting.
Contact with a capsule is not eliminated: thru scores the executed motion, and tracking error under the disturbance torque carries \methodname through a capsule on \qty{7.3}{\percent} and \qty{11.5}{\percent} of plans, against \qty{24.1}{\percent} and \qty{27.0}{\percent} for RRTC.

\subsection{Ablations}
\label{sec:ablations}

We ablate the trajectory parameterization, the probability model, and the arm dynamics model.

\emph{Trajectory parameterization.}
In our approach, each edge is a per-joint cubic, the lowest-order polynomial fixed by the four boundary conditions \((q_0, \dot q_0, q_1, \dot q_1)\). However, alternative polynomial orders exist; lower orders lack the expressiveness to represent all boundary conditions and higher orders introduce more expressivity but also additional degrees of freedom. In this section we explore the performance of alternate parameterizations. 
In \cref{tab:abl:param}, orders~4 and~5 match the cubic within a standard deviation.
Dropping the terminal velocity \(\dot q_1\) is worst, at \qty{41.7}{\percent}, as the planner can no longer command a cut speed at the rendezvous.

\input{tables/abl_param}

\emph{Scoring model.}
Arm and environment state become hard to predict over long horizons. While our model takes this into account, the removal of these terms still leave a valid planner. In \cref{tab:abl:prob} we investigate the performance after   removing terms from the objective~\eqref{eq:objective}.
\(\log P_{\mathrm{inrange}}\) dominates: without it the cut rate falls to \qty{28.3}{\percent}, as the planner picks rendezvous the object has already passed.
Removing \(\log P_{\mathrm{arm}}\) costs speed rather than contact, with \(\bar v\) falling from \qty{5.45}{} to \qty{4.38}{\metre\per\second} at unchanged catch rate, indicating that the planner commits to swings that trigger the arm's reflexes.

\input{tables/abl_prob}

\emph{Arm dynamics model.}
\cref{tab:abl:arm} removes components of the dynamics model.
Dropping any single constraint other than the velocity limit has little effect beyond the spread, while keeping only friction and armature drops the cut rate to \qty{61.1}{\percent}.

\input{tables/abl_arm}

\subsection{Real-world Hardware}
\label{sec:hardware}

We ran \methodname in the real-world with the same planner, controller, and communication stack used in simulation, and with the estimator calibrated to the hardware residuals of \cref{sec:objective}. 30 baseball-sized objects were hand-tossed into the workspace of the arm. Over the 30 tosses, \methodname executed a successful cut on 26 (86.7\unit{\percent}) and executed a successful catch on 26 (86.7\unit{\percent}). Of these, 25 successful cuts were performed consecutively. The majority of failures in the real world were due to perception errors, where thrown balls are not reliably detected by the OptiTrack system, and human error, where tosses are not as consistent as compared to simulation.

Additionally, we recorded the tossed objects in this trial and played them back in simulation, measuring the baseline planners' performance under the same distribution of tosses. These results, including those for the fully real experiments, are presented in \cref{tab:real}. Small performance differences arise compared to the fully simulated trials (\cref{tab:clutter}) because the hand-tossed objects explored a smaller portion of the arm workspace and intensified behaviors specific to those regions.

\input{tables/real}

\section{Conclusion}
\label{sec:conclusion}

We have presented \methodname, an anytime batched planner that intercepts a thrown object by growing a tree of exact cubic edges on the GPU toward an explicit chart of the interception manifold.
Each edge is timed against the arm's torque, power, and velocity limits, and plans are ranked by the probability of contact under uncertainty.
In a simulator calibrated on the physical arm, \methodname cuts \qty{96.7}{\percent} of tosses in the open and \qty{68.3}{\percent} among five obstacles, against \qty{68.3}{\percent} and \qty{35.0}{\percent} for the best single-cubic baseline; ablations show the cubic edge, the objective's arrival-time term, and the dynamics model are each necessary.
The main limitations are the ballistic flight model, which omits drag and object deformation, the arrival-time pruning rule, which is a heuristic rather than an admissible bound and so can discard a higher-scoring rendezvous, and the simple uncertainty models used for arm's own kinematics and dynamics.
Additionally, checks and the chart decoder are smooth and differentiable in the trajectory parameters, and we believe gradient-based refinement of tree nodes is a direct next step.

\ifanonymous
\section*{Acknowledgments}
    Claude was used for editing and programming assistance. All technical content and final manuscript text were reviewed and verified by the authors.
\else
    \section*{Acknowledgments}
    The authors would like to thank Jolie Lai for help manufacturing components for the real world experiments.  
    Claude was used for editing and programming assistance. All technical content and final manuscript text were reviewed and verified by the authors.
\fi

\printbibliography{}

\end{document}

%% file: preamble.tex
\usepackage[utf8]{inputenc}
\usepackage[english]{babel}

\usepackage{comment}
\usepackage{xspace}
\usepackage{graphicx}
\usepackage[dvipsnames,table]{xcolor}
\usepackage{overpic}
\usepackage{svg}
\usepackage{tikz}
\usepackage{bm}
\usepackage[hang,flushmargin]{footmisc}
\usepackage{adjustbox}
\usepackage{etoolbox}
\usepackage{environ}
\usepackage{pbalance}
\usepackage{flushend}
\usepackage{mathtools}
\usepackage[normalem]{ulem}
\usepackage{hhline}
\usepackage{tabularx}
\usepackage{multirow}
\usepackage{booktabs}
\usepackage{colortbl}
\usepackage{float}
\usepackage{placeins}
\usepackage{array}
\usepackage{threeparttable}
\usepackage{makecell}

\usepackage{amsmath}
\usepackage{amssymb}
\usepackage{amsthm}
\usepackage{siunitx}
\DeclareSIUnit[number-unit-product={}]{\percent}{\%}
\DeclareSIUnit{\frame}{frame}
\DeclareSIUnit{\swing}{swing}
\usepackage{mathtools}

\DeclareMathOperator*{\argmax}{arg\,max}
\DeclareMathOperator*{\argmin}{arg\,min}

\makeatletter
\patchcmd{\@makecaption}{\scshape}{}{}{}
\newcommand{\onetagright}{\tagsleft@false}
\makeatother

\usepackage{subcaption}
\newtheoremstyle{main}
{1em}                                                %
{1em}                                                %
{\itshape}                                           %
{0pt}                                                %
{\scshape}                                           %
{\\*}                                                %
{2pt}                                                %
{\thmname{#1}\thmnumber{ #2}: \thmnote{\itshape #3}} %

\newtheorem{definition}{Definition}[section]

\usepackage[linesnumbered,ruled,noend]{algorithm2e}
\makeatletter
\renewcommand{\algocf@linesnumbered}{%
  \algocf@seteveryparnl{\refstepcounter{AlgoLine}\gdef\@currentlabel{\theAlgoLine}}}
\makeatother
\makeatletter
\newcommand{\removelatexerror}{\let\@latex@error\@gobble}
\makeatother

\let\labelindent\relax
\usepackage[inline]{enumitem}

\usepackage[
  activate   = {true},
  protrusion = false,
  expansion  = true,
  kerning    = true,
  spacing    = true,
  tracking   = false,
  auto       = true,
  selected   = true,
  factor     = 1000,
  stretch    = 10,
  shrink     = 10,
]{microtype}

\usepackage{csquotes}
\usepackage[
  maxbibnames=6,
  minbibnames=1,
  maxcitenames=2,
  natbib=true,
  bibstyle=ieee,
  citestyle=numeric-comp,
  backend=biber,
  sorting=none,
  giveninits=true,
  url=false,
  doi=false,
  eprint=false,
  isbn=false,
]{biblatex}

\definecolor{purduegold}{HTML}{000000} %

\makeatletter
\let\NAT@parse\undefined
\makeatother
\usepackage[pdfa,colorlinks,bookmarksopen,bookmarksnumbered,allcolors=purduegold]{hyperref}
\usepackage{bookmark}

\usepackage[nameinlink,capitalise]{cleveref}
\crefname{line}{line}{lines}
\crefname{AlgoLine}{line}{lines}
\Crefname{AlgoLine}{Line}{Lines}
\crefname{figure}{Fig.}{Figs.}
\Crefname{figure}{Fig.}{Figs.}
\crefname{equation}{Eq.}{Eqs.}
\Crefname{equation}{Eq.}{Eqs.}
\crefname{section}{Sec.}{Secs.}
\Crefname{section}{Sec.}{Secs.}
\crefname{definition}{Def.}{Defs.}
\Crefname{definition}{Def.}{Defs.}
\crefname{algorithm}{Alg.}{Algs.}
\Crefname{algorithm}{Alg.}{Algs.}
\crefname{assumption}{Asm.}{Asms.}
\Crefname{assumption}{Asm.}{Asms.}
\crefname{subassumption}{Asm.}{Asms.}
\Crefname{subassumption}{Asm.}{Asms.}
\crefname{problem}{Problem}{Problems}
\Crefname{problem}{Problem}{Problems}

\makeatletter
\newcommand\footnoteref[1]{\protected@xdef\@thefnmark{\ref{#1}}\@footnotemark}
\makeatother

\graphicspath{ {./figures/} }

\definecolor{goldbest}{HTML}{F3C969}
\newcolumntype{R}{>{\raggedleft\arraybackslash}X}

\newcommand{\R}{\mathbb{R}}

\newcommand{\tfall}{t_{\mathrm{fall}}}

\newcommand{\Mman}{\mathcal{M}}

%% file: commands.tex
\providecommand{\best}[1]{\textbf{#1}}

\newcommand{\methodname}{FRUITNINJA\xspace}

%% file: big_block_of_tables.tex
\begin{figure*}[!ht]
  \centering
  \makebox[\linewidth][c]{\includegraphics[width=1.05\linewidth]{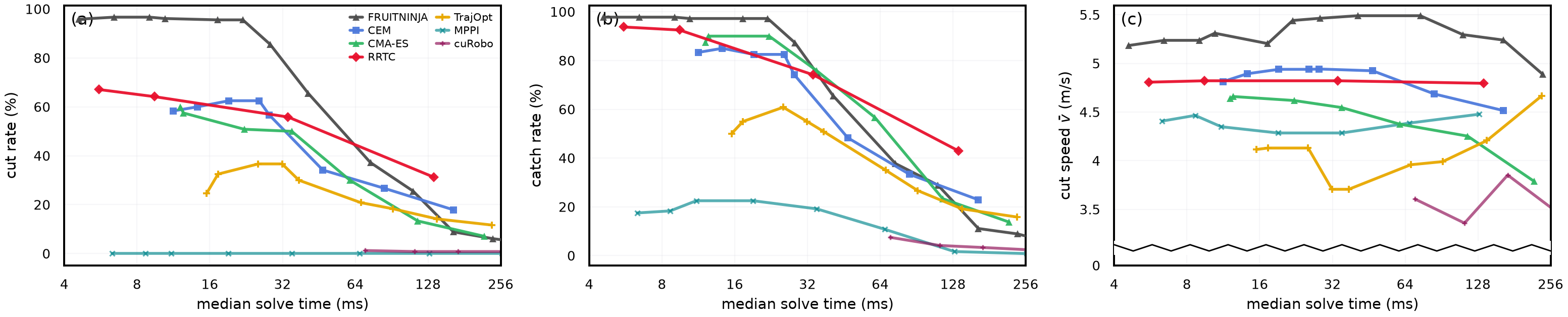}}
  \caption{Cut rate, catch rate, and cut speed vs.\ median solve time. Each point is one planner configuration (varying batch size, iterations, or population). \methodname reaches \qty{96.7}{\percent} cut rate at \qty{12}{\milli\second}; baselines plateau below \qty{70}{\percent}.}
  \label{fig:new_fig_16_frontier}
\end{figure*}

\begin{table*}[!ht]
  \centering

  \begin{tabular*}{\textwidth}{@{\extracolsep{\fill}}l *{3}{S[table-format=2.1] S[table-format=2.1] S[table-format=2.1] S[table-format=1.2]}@{}}
    & \multicolumn{4}{c}{\textbf{\(N=0\)} (no obstacle)} & \multicolumn{4}{c}{\textbf{\(N=3\)} (3 capsules)} & \multicolumn{4}{c}{\textbf{\(N=5\)} (5 capsules)} \\
    \cmidrule(lr){2-5} \cmidrule(lr){6-9} \cmidrule(lr){10-13}
    planner & cut\(\uparrow\) & catch\(\uparrow\) & thru\(\downarrow\) & \(\bar v\uparrow\) & cut\(\uparrow\) & catch\(\uparrow\) & thru\(\downarrow\) & \(\bar v\uparrow\) & cut\(\uparrow\) & catch\(\uparrow\) & thru\(\downarrow\) & \(\bar v\uparrow\) \\
    \midrule
    \textbf{\methodname (ours)} & \cellcolor{goldbest}\textbf{96.7} & \textbf{98.3} & \(\times\) & \textbf{5.35} & \cellcolor{goldbest}\textbf{81.7} & \textbf{85.0} & 7.3 & 5.36 & \cellcolor{goldbest}\textbf{68.3} & \textbf{81.7} & 11.5 & \textbf{5.06} \\
    RRTC (CPU) & 73.3 & \textbf{98.3} & \(\times\) & 4.76 & 58.3 & 78.3 & 24.1 & 4.75 & 21.7 & 36.7 & 27.0 & 4.27 \\
    CEM & 68.3 & 76.7 & \(\times\) & 5.30 & 53.3 & 65.0 & 4.7 & \textbf{5.37} & 38.3 & 53.3 & \textbf{5.3} & 5.03 \\
    CMA-ES & 68.3 & 86.7 & \(\times\) & 4.96 & 65.0 & 80.0 & \textbf{5.0} & 5.11 & 35.0 & 68.3 & 5.4 & 5.00 \\
    TrajOpt & 41.7 & 73.3 & \(\times\) & 3.81 & 31.7 & 51.7 & 8.1 & 3.80 & 43.3 & 63.3 & 14.5 & 3.79 \\
    MPPI & 0.0 & 30.0 & \(\times\) & \(\times\) & 0.0 & 15.0 & 7.7 & \(\times\) & 0.0 & 21.7 & 16.3 & \(\times\) \\
    cuRobo & 1.7 & 6.7 & \(\times\) & 2.38 & 1.7 & 5.0 & 9.7 & 4.18 & 0.0 & 3.3 & 10.9 & \(\times\) \\
    \bottomrule
  \end{tabular*}

  \caption{Obstacle avoidance (60 tosses per clutter level). Cut/catch: \unit{\percent}; \(\bar v\): contact-conditional speed (\unit{\metre\per\second}); thru: fraction of plans whose executed motion passes through a capsule. \colorbox{goldbest}{Gold}: best cut rate per level. Averages are calculated only on successful trials. \methodname maintains the best cut and catch rate even as obstacle density increases.}
  \label{tab:clutter}

  \vspace{-1.5em}
  
\end{table*}

%% file: tables/abl_param.tex
\providecommand{\best}[1]{\textbf{#1}}
\providecommand{\pend}{\textcolor{gray}{--}}
\begin{table}[t]
\centering\small
\begin{tabular*}{\columnwidth}{@{\extracolsep{\fill}}lccc@{}}
variant & cut (\%)$\uparrow$ & $\bar v$ (m/s)$\uparrow$ & solve (ms)$\downarrow$ \\
\midrule
cubic (control) & \best{$96.1 \pm 1.0$} & $5.41 \pm 0.06$ & $9.65 \pm 0.07$ \\
\midrule
order 1 & $78.9 \pm 7.5$ & \bm{$6.13 \pm 0.09$} & \bm{$8.26 \pm 0.12$} \\
order 2, no $\dot q_1$ & $41.7 \pm 2.9$ & $4.74 \pm 0.11$ & $8.40 \pm 0.07$ \\
order 2, no $\dot q_0$ & $76.7 \pm 4.4$ & $5.25 \pm 0.02$ & $9.02 \pm 0.15$ \\
order 4 & $94.3 \pm 0.9$ & $5.47 \pm 0.07$ & $9.10 \pm 0.30$ \\
order 5 (min $\dddot q$) & \bm{$95.6 \pm 1.0$} & $5.41 \pm 0.07$ & $9.25 \pm 0.18$ \\
task-space cubic & $13.3 \pm 2.9$ & $4.56 \pm 0.11$ & $30.80 \pm 0.07$ \\
\bottomrule
\end{tabular*}
\caption{Trajectory parameterization. Each variant is three seeds $\times$ $60$ throws.}
\label{tab:abl:param}

\vspace{-1em}

\end{table}

%% file: tables/abl_prob.tex
\providecommand{\best}[1]{\textbf{#1}}
\providecommand{\pend}{\textcolor{gray}{--}}
\begin{table}[t]
\centering\small
\setlength{\tabcolsep}{3pt}
\begin{tabular*}{\columnwidth}{@{\extracolsep{\fill}}lccc@{}}
ablation & cut (\%)$\uparrow$ & catch (\%)$\uparrow$ & $\bar v$ (m/s)$\uparrow$ \\
\midrule
full objective (control)           & $96.7 \pm 1.7$  & $97.2 \pm 2.0$  & $5.45 \pm 0.07$ \\
\midrule
$-\log P_\text{object}$   & \bm{$90.0 \pm 2.3$}  & $94.2 \pm 1.2$  & \bm{$5.33 \pm 0.00$} \\
$-\log P_\text{inrange}$  & $28.3 \pm 10.3$  & $47.8 \pm 9.7$  & $4.96 \pm 0.28$ \\
$-\log P_\text{arm}$      & $85.0 \pm 0.0$  & \bm{$96.7 \pm 0.0$}  & $4.38 \pm 0.01$ \\
\midrule
flat covariance          & $85.0 \pm 2.3$  & $93.3 \pm 2.3$  & $5.45 \pm 0.00$ \\
\bottomrule
\end{tabular*}
\caption{Probability model ablation. Each row removes one or two terms from the objective~\eqref{eq:objective}; the pairwise rows remove both named terms. ``Flat covariance'' freezes $\sigma(t)$ at its initial value. Mean $\pm$ std.\ dev.\ across seeds.}
\label{tab:abl:prob}

\vspace{-2em}
\end{table}

%% file: tables/abl_arm.tex
\providecommand{\best}[1]{\textbf{#1}}
\providecommand{\pend}{\textcolor{gray}{--}}
\begin{table}[t]
  \centering\small
  \begin{tabular*}{\columnwidth}{@{\extracolsep{\fill}}lccc@{}}
    variant & cut (\%)\(\uparrow\) & catch (\%)\(\uparrow\) & \(\bar v\) (m/s)\(\uparrow\) \\
    \midrule
    control                & \(97.2 \pm 1.0\) & \(98.3 \pm 0.0\) & \(5.37 \pm 0.06\) \\
    \midrule
    \multicolumn{4}{l}{\emph{leave one out}}\\
    \(-\)torque              & \(\bm{96.7 \pm 1.7}\) & \(97.8 \pm 1.0\) & \(\bm{5.37 \pm 0.06}\) \\
    \(-\)velocity            & \(84.4 \pm 5.1\) & \(94.4 \pm 2.5\) & \(5.33 \pm 0.09\) \\
    \(-\)friction, armature  & \(\bm{96.7 \pm 1.7}\) & \(\bm{98.3 \pm 0.0}\) & \(5.31 \pm 0.06\) \\
    \midrule
    \multicolumn{4}{l}{\emph{leave one in}}\\
    \(+\)torque              & \(\bm{87.8 \pm 1.0}\) & \(95.0 \pm 2.9\) & \(\bm{5.42 \pm 0.05}\) \\
    \(+\)velocity            & \(87.8 \pm 3.5\) & \(\bm{96.7 \pm 2.9}\) & \(5.36 \pm 0.06\) \\
    \(+\)friction, armature  & \(61.1 \pm 4.2\) & \(72.2 \pm 1.9\) & \(5.29 \pm 0.09\) \\
    \midrule
    nothing modelled       & \(65.6 \pm 6.3\) & \(71.7 \pm 4.4\) & \(5.38 \pm 0.12\) \\
    \bottomrule
  \end{tabular*}
  \caption{Arm dynamics model ablation (three seeds \(\times\) 60 throws per variant). Velocity removes the joint-velocity ratio and position-dependent taper; friction and armature are \cref{eq:torque}.}
  \label{tab:abl:arm}

  \vspace{-0.5em}
  
\end{table}

%% file: tables/real.tex
\providecommand{\methodname}{FRUITNINJA}
\providecommand{\best}[1]{\textbf{#1}}
\begin{table}[t]
\centering\small
\begin{tabular*}{\columnwidth}{@{\extracolsep{\fill}}lccc@{}}
 & cut\(\uparrow\) & catch\(\uparrow\) & \(\bar v\uparrow\) \\
\midrule
real control & 86.7 \(\pm\) 6.2 & 86.7 \(\pm\) 6.2 & 4.29 \(\pm\) 0.20 \\
\midrule
\multicolumn{4}{l}{\emph{simulation}}\\
\textbf{\methodname (ours)} & \cellcolor{goldbest}\textbf{86.7 \(\pm\) 6.2} & \textbf{86.7 \(\pm\) 6.2} & 4.93 \(\pm\) 0.13 \\
RRTC (CPU) & 50.0 \(\pm\) 9.1 & 73.3 \(\pm\) 8.1 & 4.63 \(\pm\) 0.37 \\
CEM & 70.0 \(\pm\) 8.4 & 76.7 \(\pm\) 7.7 & \textbf{5.42 \(\pm\) 0.18} \\
CMA-ES & 33.3 \(\pm\) 8.6 & 46.7 \(\pm\) 9.1 & 4.56 \(\pm\) 0.32 \\
TrajOpt & 66.7 \(\pm\) 8.6 & 70.0 \(\pm\) 8.4 & 3.33 \(\pm\) 0.16 \\
MPPI & 20.0 \(\pm\) 7.3 & 30.0 \(\pm\) 8.4 & 4.42 \(\pm\) 0.22 \\
cuRobo & 6.7 \(\pm\) 4.6 & 10.0 \(\pm\) 5.5 & 4.46 \(\pm\) 0.15 \\
\bottomrule
\end{tabular*}
\caption{Real object trajectories with simulated FR3. The ``real control" row is a trial done fully in the real world.}
\label{tab:real}

\vspace{-1.5em}
\end{table}